\documentclass[10pt]{article}
\usepackage{blindtext}
\usepackage[a4paper, total={6in, 9in}]{geometry}
\usepackage{graphicx}
\usepackage{listings}
\usepackage{indentfirst,csquotes}
\usepackage{amsmath}
\usepackage{xcolor,paralist,hyperref,titlesec,fancyhdr,etoolbox}
\usepackage{pdflscape}
\usepackage{amsthm,amssymb}
\usepackage{authblk}
\usepackage{rotating}
\usepackage{tocloft}

\newtheorem{theorem}{Theorem}[]
\newtheorem{definition}[theorem]{Definition}

\titlespacing*{\section}{0pt}{1ex plus 0.5ex minus 0.2ex}{0.5ex plus 0.2ex minus 0.2ex}

\hypersetup{colorlinks=true, linkcolor=black, filecolor=black, urlcolor=black}

\cftsetindents{section}{1em}{1em}
\newcommand{\listequationsname}{\fontsize{12}{14}\selectfont List of Equations} 
\newlistof{myequations}{equ}{\listequationsname}
\newcommand{\myequations}[1]{%
\addcontentsline{equ}{myequations}{\protect\numberline{\theequation}#1}\par
}

\title{Need of AI in Modern Education: in the Eyes of Explainable AI (xAI)}
\author{Supriya Manna, Niladri Sett\footnote{all corresponds to: \textit{settniladri@gmail.com}}}
\affil{SRM University AP \\ Amaravati, Andhra Pradesh 522502, India}
\date{}
\begin{document}

    
    
    
    

\maketitle




\section{Abstract}
     Modern Education is not \textit{Modern} without AI. However, AI's complex nature makes understanding and fixing problems challenging. Research worldwide shows that a parent's income greatly influences a child's education. This led us to explore how AI, especially complex models, makes important decisions using Explainable AI tools. Our research uncovered many complexities linked to parental income and offered reasonable explanations for these decisions. However, we also found biases in AI that go against what we want from AI in education: clear transparency and equal access for everyone. These biases can impact families and children's schooling, highlighting the need for better AI solutions that offer fair opportunities to all. This chapter tries to shed light on the complex ways AI operates, especially concerning biases. These are the foundational steps towards better educational policies, which include using AI in ways that are more reliable, accountable, and beneficial for everyone involved.

\section{Introduction}
\let\thefootnote\relax
\footnotetext{
codes: \href{https://colab.research.google.com/drive/149rwtHbcBqbKFuNWbXz7PhAmV_cdSpzW?usp=sharing.}{notebook}}
The trajectory of the education system has traversed a remarkable journey from medieval times to the modern era, reflecting societal advancements, technological innovations, and evolving pedagogical paradigms. Initially characterized by rudimentary pedagogical methods, limited accessibility, and a rigid curriculum, the education landscape has undergone transformative changes, albeit with persistent challenges.

Historically, the medieval education system was predominantly characterized by religious institutions, emphasizing classical literature, theology, and rote memorization. Access was largely restricted to elite echelons of society, with pedagogical approaches rooted in tradition and conformity.

Fast-forwarding to the modern era, the education system has witnessed democratization, technological integration, and a paradigm shift towards personalized learning. However, contemporary challenges persist, necessitating strategic interventions:

\begin{itemize}
    \item\textbf{Better Recommendation of Courses:}
Traditional systems often adopt a one-size-fits-all approach, overlooking individual learning preferences and aptitudes. AI-powered recommender systems being employed in book selection  \cite{ai recom}, utilizing collaborative filtering algorithms, analysing student data to generate personalized course recommendations, and optimizing learning outcomes and engagement. Contemporary software solutions, such as \textit{Coursera}'s recommendation engine\textsuperscript{1}\footnote{\href{https://www.coursera.org/}{1. https://www.coursera.org/}}.
 and platforms like \textit{Knewton Alta} \textsuperscript{2}\footnote{\href{https://www.wiley.com/en-us/education/alta}{2. https://www.wiley.com/en-us/education/alta}}
 leverage advanced algorithms to tailor learning pathways based on individual student profiles.

\item\textbf{SWOT Analysis of Students:}
Comprehensive student evaluation transcends academic performance, necessitating a nuanced understanding of individual strengths, weaknesses, opportunities, and threats. Machine learning models, employing clustering analysis algorithms  \cite{clus algo}, facilitate intricate SWOT analyses, enabling educators to devise tailored learning strategies and interventions. Software applications like \textit{IBM Watson}Analytics\textsuperscript{3}\footnote{\href{https://www.ibm.com/watson}{3. https://www.ibm.com/watson}}
 and \textit{Tableau}\textsuperscript{4}\footnote{\href{https://www.tableau.com/}{4. https://www.tableau.com/}}
 harness sophisticated algorithms to decode complex student profiles, fostering personalized academic trajectories.

\item\textbf{Transparency for Students:}
The ambiguity surrounding grading criteria, learning objectives, and assessment methodologies undermines student engagement and comprehension. Natural Language Processing (NLP) techniques, coupled with sentiment analysis algorithms  \cite{senti}, elucidate grading metrics and curricular objectives, fostering transparent and collaborative learning environments. Modern platforms like \textit{Turnitin}\textsuperscript{5}\footnote{\href{https://www.turnitin.com/}{5. https://www.turnitin.com/}}
 and \textit{Grammarly}\textsuperscript{6}\footnote{\href{https://www.grammarly.com/}{6. https://www.grammarly.com/}}
 employ AI-driven NLP tools to provide actionable feedback, enhancing transparency and academic integrity.

\item\textbf{Better Performance Indicators with Personalized Feedback:}
Conventional performance metrics often lack granularity, overshadowing individual accomplishments and areas of improvement. AI-driven data analytics platforms, harnessing predictive analytics algorithms, empower educators to deliver personalized performance indicators and actionable insights, optimizing learning trajectories and student engagement. Software solutions such as \textit{Blackboard Analytics}\textsuperscript{7}\footnote{\href{https://www.blackboard.com}{7. https://www.blackboard.com}} and \textit{Moodle}\textsuperscript{8}\footnote{\href{https://moodle.org/}{8. https://moodle.org/}} leverage advanced algorithms to analyze student data, facilitating targeted interventions and pedagogical enhancements.

\end{itemize}

\subsection{  The Point of This Book Chapter:}

Numerous studies have underscored the profound potential of Artificial Intelligence (AI) in augmenting and revitalizing the contemporary education system  \cite{edu1, edu2, edu3, edu4}. However, the integration of 'Modern' education with AI is not devoid of challenges, particularly concerning financial implications and accessibility disparities.

A plethora of global research, spanning developed nations such as Norway  \cite{study3}, Japan  \cite{study1}, developing nations such as China  \cite{study2} to least developed African countries  \cite{study4} like Ghana  \cite{study5}, elucidates a direct correlation between parental income and the quality of education availed by their progeny. In the 21st century landscape, while AI embodies an indispensable cornerstone of educational advancement, its implementation entails substantial costs encompassing resource allocation, hiring, cross-testing, verification, licensing, and deployment  \cite{cost of ai}. Consequently, the escalating financial exigencies associated with \textit{modern} education exacerbate accessibility barriers\textsuperscript{9,}\footnote{\href{https://www.thehindubusinessline.com/opinion/columns/c-p-chandrasekhar/the-alarming-rise-in-education-costs-in-new-india/article33215181.ece}{9. https://www.thehindubusinessline.com/opinion/columns/c-p-chandrasekhar/the-alarming-rise-in-education-costs-in-new-india/article33215181.ece}}
\textsuperscript{10}\footnote{\href{https://timesofindia.indiatimes.com/india/rising-cost-of-education-worries-parents-survey-shows/articleshow/17946981.cms?from=mdr}{10. https://timesofindia.indiatimes.com/india/rising-cost-of-education-worries-parents-survey-shows/articleshow/17946981.cms?from=mdr}}
 \cite{cost of higher education}, constraining students hailing from economically disadvantaged backgrounds. Regrettably, instances abound wherein academically meritorious students, despite securing admissions into prestigious global universities, confront insurmountable financial hurdles, precluding educational attainment due to parental income constraints.

We acknowledge the fact that parental income substantially influences the attainability of educational opportunities for students. In this study, we want to investigate this aspect of parental income. How the income of an individual is dependent on which factors. How can we visualise those dependencies? Moreover, are those dependencies fair? If not, how can we investigate the unfairness? Is the Notion of \textit{feature importance} enough to recognise a biased model? This book chapter endeavours to delve deeper into these pivotal nexus. We use several methodologies for visualising the feature importance from a model and then we investigate whether are they enough for investigating (un)fairness. We use the adult census dataset in this study to work on this binary classification problem of predicting whether an individual (can) earn more than 50k USD a year, given the other attribute.  This is the primary indicator for our study where we're presuming (based on the earlier studies mentioned above) that the individual, as a \textit{parent}, enhances the chance of availing \textit{better} education for his progeny. Through rigorous analysis and interpretation, this chapter aspires to furnish insights, foster dialogue, and catalyze informed interventions addressing the confluence of AI, education, and socio-economic disparities, steering towards an inclusive and equitable educational landscape. 

In this book chapter, we have extensively scrutinized several recent research papers, tutorials, official documentation, and internet articles to gather information and reproduce the relevant results from current research. This book chapter is meant to be suitable for both technical as well as non-technical readers. We have, furthermore, considered numerous ways of structuring this chapter inspired by our scrutinization. The links to relevant \textit{resources} are kept in the footnotes below. 

Finally, Our study, by no means, is meant to be devised as an Oracle but we fundamentally believe in asking for explanations for high-stakes decisions taken by AI models for the sake of trustworthiness. In the education sector, this has been stressed in recent times \cite{trust in education}, and in this chapter, we are demonstrating the same on one of the fundamental aspects of availing 'Modern' education: parental income using state-of-the-art xAI techniques.
\footnote{\textit{resources}:}
\footnote{\url{https://bit.ly/3u2v0ay}}
\footnote{\url{https://rb.gy/dd2dcx}}
\footnote{\url{https://bit.ly/3S1S9lo}}
\footnote{\url{https://rb.gy/ekghfm}}
\footnote{\url{https://rb.gy/qusxjj}}
\footnote{\url{https://rb.gy/fq4ysa}}
\footnote{\url{https://shorturl.at/dhAO7}}
\footnote{\url{https://shorturl.at/ntvBH}}
\footnote{\url{https://bit.ly/425kmg4}}
\footnote{\url{https://bit.ly/3O2DBko}}
\footnote{\url{https://bit.ly/3SjuPRR}}

\section{Inherent Complexity: A Double-Edged Sword}

As discussed, while AI harbours immense potential to revolutionize education, fostering personalized learning, enhancing administrative efficiency, and facilitating data-driven decision-making, its inherent complexity poses formidable challenges, particularly concerning transparency and interpretability in deducing decisions which concurrently introduces nuanced challenges that necessitate vigilant scrutiny and strategic navigation. Sophisticated AI software including Llms \cite{llama, gemini, gpt}, SaaS and commercial software such as \cite{h2o, einstein}, and virtual assistants \cite{alexa et all}, while endowed with remarkable capabilities, often manifests as convoluted enigmas, presenting two underlying problems:

\begin{enumerate}
    \item \textbf{Explanation Gap}: AI models, characterized by intricate hyperspaces defined by multifaceted weights and attributes, inherently lack traceability and debuggability. The stochastic nature of the training phase, coupled with the intricacies of the hyperspace, obfuscates the retrieval and comprehension of model parameters and rationales. Consequently, elucidating the \textit{'why'} behind AI predictions remains a daunting endeavor, devoid of straightforward exploration avenues.
    
    \item \textbf{Trust in the \textit{`Black-Box'}}: For seasoned machine learning practitioners well-versed in model architectures and training phases, navigating the intricacies of AI may seem feasible. However, for end-users devoid of specialized expertise, AI often emerges as an opaque 'black box', necessitating dual proficiency as AI and domain experts to engender trust. While domain expertise may be commonplace, it scarcely equates to an unwavering confidence in results generated by intricate AI systems, perpetuating scepticism and apprehension.
\end{enumerate}
\subsection{Explainable AI (XAI): Bridging the Trust Gap}

In the pursuit of deciphering the intricacies of 'black-box' AI systems and fostering trust among end-users, scientists have embarked on a relentless quest, culminating in the emergence of Explainable AI (XAI) as an active area of research in this decade. XAI transcends traditional AI paradigms, endeavouring to illuminate the opaque nature of complex machine learning models by enabling:

\begin{enumerate}
    \item \textbf{Enhanced Explainability}: XAI endeavours to cultivate machine learning techniques that aim to yield more explainable models without compromising learning performance or prediction accuracy \cite{xai def}. By elucidating model rationale and inherent strengths and weaknesses, XAI augments user comprehension and trust in several high stakes (for example: \cite{xai trust}), pivotal for fostering collaborative partnerships between humans and AI entities.
    
    \item \textbf{Human-Centric Design}: XAI advocates for the integration of state-of-the-art human-computer interface techniques, facilitating the translation of intricate models into comprehensible and actionable insights for end-users \cite{hai}. By crafting intuitive explanation dialogues and interfaces, XAI empowers users to navigate, interpret, and leverage AI capabilities effectively, transcending the complexities of underlying algorithms and architectures.
    
    \item \textbf{Diverse Methodological Portfolio}: Recognizing the multifaceted challenges inherent in achieving explainability without compromising performance, XAI adopts a multifaceted approach, exploring and integrating diverse techniques across the performance-versus-explainability trade space \cite{perf vs exp}. By curating a portfolio of methodologies, XAI equips developers with versatile design options, facilitating tailored solutions aligned with specific application domains and user requirements.
\end{enumerate}

\section{Methodology: Explainability, Interpretability and more}
Before jumping into the technical and mathematical intricacies of xAI, it is critical to distinguish between explainability and interpretability in machine learning applications for our objectives. We regard interpretability as a means of obtaining explainability. Explainable AI (or XAI) refers to models whose results can be comprehended by ordinary humans. However, this does not imply that the model is required to be interpretable. That is, there are two types of models: interpretable models and non-interpretable models. \textit{Linear regression, logistic regression, decision trees} etc are examples of interpretable models. Practitioners can simply determine how the model outputs either its label or value. Then there are non-interpretable models for image processing and natural language processing, such as random forests, feed-forward neural networks, and deep learning architectures \cite{xAI Book}.
In order to make sense of the process by which the intricate model that was employed to generate the forecast came to be, we employ post hoc explanations \cite{post} for these. This is frequently accomplished by examining the model's gradients or using stand-in models to simulate behavior in specific areas.

We believe that the line that separates interpretability from explainability also follows the processes of the human mind. For example, human beings occasionally use deliberate, logical thinking to arrive at a conclusion. As a result, the logic can be examined in detail.
In other cases, people base their initial conclusions on intuition after analyzing complex facts over extended periods of time. Then, just as we use post hoc explanations on black box models, the people look for an explanation for a decision. Determining which strategy is superior for explainable AI is probably going to need major advancements in human cognitive science and our comprehension of the functioning of the human brain.
\subsection{Post-hoc Explainations}
In regards to the realistic scenario \cite{post}, in which the end user looks at the details after a study has been concluded and the data collected, our goal is to provide context for the predictions of a machine learning model. In order to do this, we must first define what it means to produce an explanation. An explanation typically states that some features are more significant than others or explains how an input feature can influence the output's magnitude in a positive or negative way.  It usually links the feature values of an instance to its model prediction in a way that is easily understood by humans. Not all explanations must be focused on the significance of a feature. They could also be particular cases or groups of influential instances. They could be real English languages or collections of rules. In fact, difficulties with homogeneous judgment arise from a variety of reasons. In our current study, we focus primarily on the importance of the feature.

Post hoc approaches might be model-agnostic or model-dependent, global or local.
Model-dependent interpretation techniques are focused on a particular model or set of models. Model-agnostic tools, on the other hand, can be used on any machine learning model, no matter how complex. These agnostic approaches typically analyze feature input and output pairs. Model-agnostic methods frequently profit from the reality that certain practitioners may not have access to the original model's specific weights and parameters. In our current work, we have been focused on local `Post-hoc' explanation methods in model-agnostic fashion. Below is the description of the methods we used in study for generating post-hoc explanation.
\begin{enumerate}
    \item \textbf{LIME} \cite{LIME}:
One of the most well-known post hoc methods is LIME or local interpretable model-agnostic explanations. LIME's concept is to build a surrogate model in a local area at a certain target point to explain the significance of the input features. Because it hypothesized that sophisticated black boxes tend to demonstrate more linear or simple behavior in local neighborhoods. We build new samples for LIME by perturbing for the target sample. We then query the black-bo model to obtain the label of the perturbed data points, scores them using a kernel, and train a \textit{sparse linear} model using the data-points and labels.
\item \textbf{KernalSHAP} \cite{SHAP}:
	Shapley values \cite{SHAPLY values} are based on the idea that a prediction may be explained by imagining each feature value of the instance as a 'player' in a game where the prediction is the payoff. Shapley values, a strategy from coalitional game theory, teach us how to allocate the "payout" among the characteristics in a \textit{fair} manner.

Shapley regression values serve as indicators of feature importance in linear models, particularly when confronted with multicollinearity. This approach mandates the model's retraining across all conceivable feature subsets $S \subseteq F$, where $F$ denotes the complete set of features. Each feature is assigned an importance value, signifying the impact of its inclusion on the model prediction. The computation involves training a model $f_{S \cup\{i\}}$ with the specific feature and another model $f_S$ without it. Subsequently, the discrepancy in the predictions for the current input is assessed as $f_{S \cup\{i\}}\left(x_{S \cup\{i\}}\right)-f_S\left(x_S\right)$, where $x_S$ represents the input feature values within the set $S$. As the influence of excluding a feature is contingent on other model features, these differences are computed for all feasible subsets $S \subseteq F \backslash\{i\}$. Subsequently, the Shapley values are calculated and used as feature attributions, representing a weighted average of all potential differences \cite{SHAPLY values, SHAP}:

\begin{equation}
\phi_i = \sum_{S \subseteq F \backslash\{i\}} \frac{|S| !(|F|-|S|-1) !}{|F| !} \left[f_{S \cup\{i\}}\left(x_{S \cup\{i\}}\right) - f_S\left(x_S\right)\right]
\end{equation}
\myequations{shaply value calculation}

However, both LIME and Kernel SHAP share the same foundation but choose different kernels and loss functions to calculate the importance of features \cite{closed form solution, all exp in one}.
With this, we are now moving to \textit{Definitions}. Thoughts have been mainly drawn from  \cite{def1, def2, def3}
\begin{definition}[Algorithm Interpretability]
An algorithm's interpretability is its ability to give users enough expressive data to comprehend how the algorithm operates. In this case, human-readable text or graphics may be considered interpretable domains. "If something is interpretable, it is possible to find its meaning or to find a particular meaning in it," according to the Cambridge Dictionary.
\end{definition}

\begin{definition}[Interpretation]
Interpretation is the process of reducing a complicated domain—like the outputs of a machine learning model—to meaningful, rational, and human-understandable ideas.
\end{definition}

\begin{definition}[Explanation]
An explanation is extra metadata that describes the feature importance or relevance of an input instance to a certain output classification. It can be produced by the machine learning model itself or by an external method.
\end{definition}

\end{enumerate}

An xAI post hoc explanation (here, LIME \& KernelShaP) works by assuming that the model architecture is a black box.  Most post hoc explainers build a single loss function using heuristics, gradients, game theory, or some other method \cite{all exp in one}. However, the black-box functions are based on the feature set's global domain. These explainers, on the other hand, mimic functions in the local domain perturbing a few examples.
    
\section{Experimental Setup}
\begin{figure}*
  \includegraphics[width=0.5\textwidth]{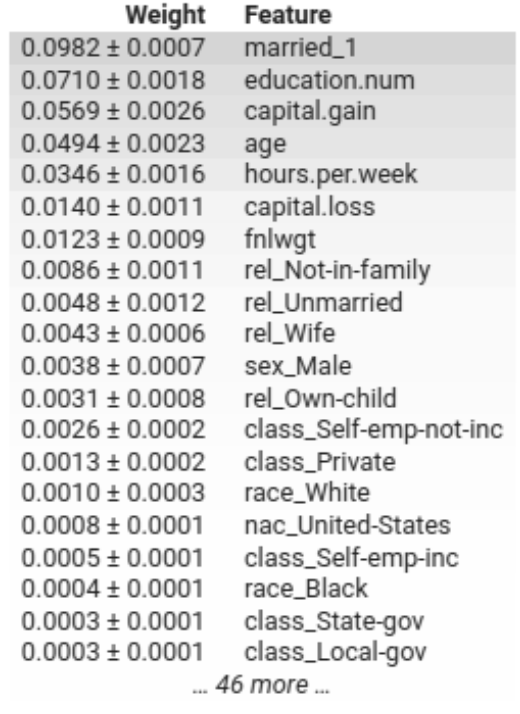}
  \includegraphics[width=0.5\textwidth]{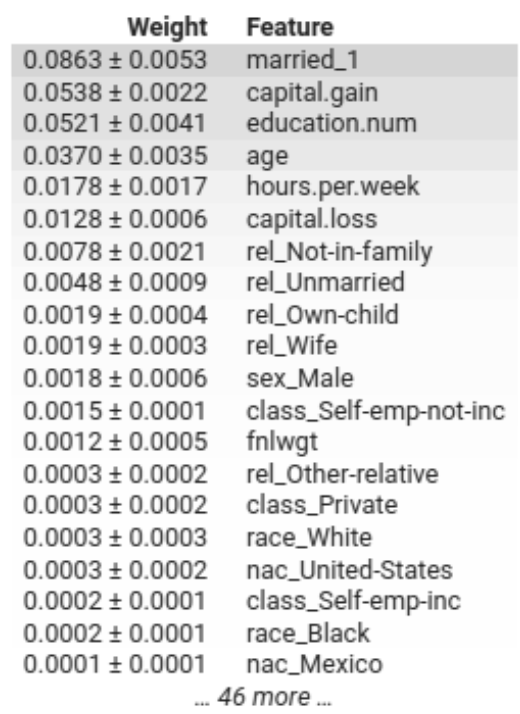}
  \caption{permutation importance}
  \label{Figure 4.}
\end{figure}

\begin{figure}
  \includegraphics[width=\textwidth]{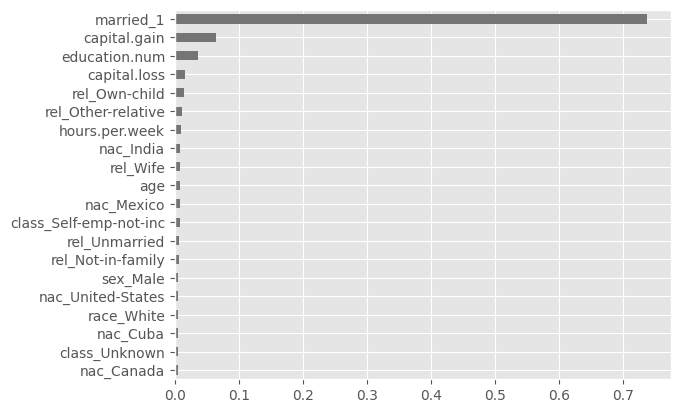}
  \caption{XGB feature importance}
  \label{Figure 1.}
\end{figure}

\begin{figure}
  \includegraphics[width=\textwidth]{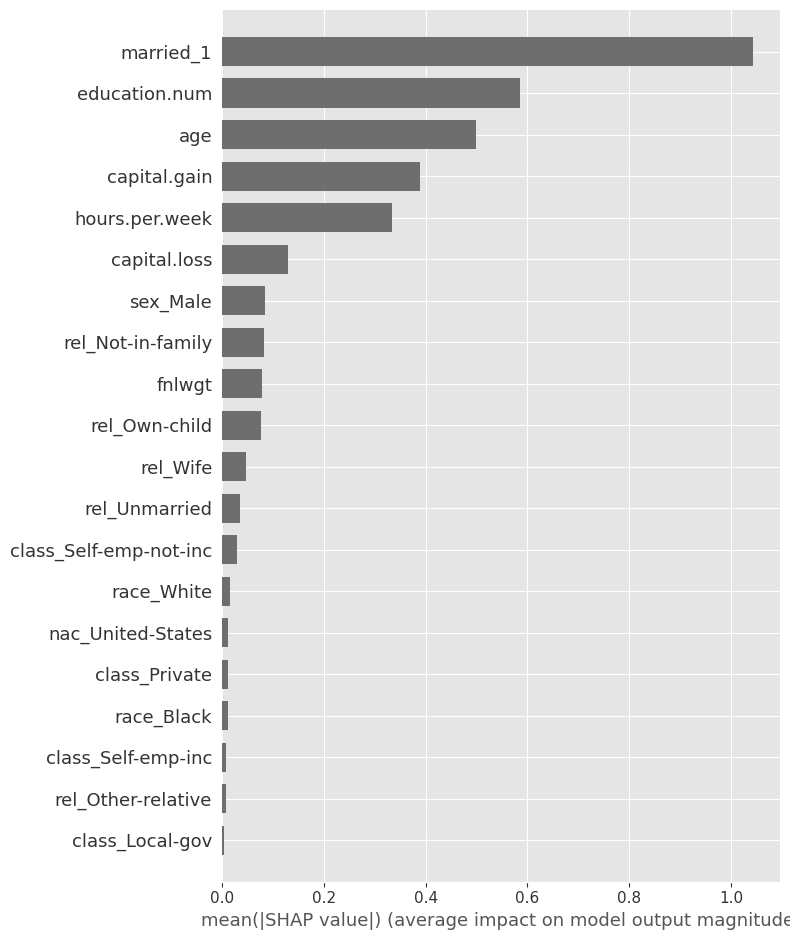}
  \caption{SHAP feature importance}
  \label{Figure 2.}
\end{figure}

The study follows the setup given in \cite{article-1,article-2} for experimentation and implementation particulars. It uses the \textit{Adult Census Income} dataset from the \textit{UCI Machine Learning Repository}. We start our experiments by training an \textit{XGBoost} \cite{xgb}  model.
The following performance metrics were achieved for the model:\newline
\begin{tabular}{rrrrrr} 
& Accuracy & Precision & Recall & F1 & AUC \\
\hline Train & 0.88 & 0.82& 0.65& 0.73 & 0.95\\
Test & 0.87 & 0.78& 0.62& 0.69& 0.92\end{tabular}
\newline
We start our investigation on the trained XGBoost based on easy to complex hypothesis gradually for calculating feature importance. We start with the ELI5 library\cite{el5}. We used the library's `Permutation Importance' (PI) method to compute the feature importance. It works on the hypothesis that the importance of a feature can be quantified by the drop in the intended metrics (accuracy, precision etc) when removed. 
In this study, we use AUC \cite{AUC} as the performance metric to evaluate the trained \textit{XGBoost} model. We report the PI for both train and test counterparts in Figure \ref{Figure 4.}: the left column shows the PI in the train set and the one in the right is for the test.
Despite the fact that the order of the most significant traits varies, it appears that the most crucial one (\texttt{married\_1}) remains the same in both folds. Furthermore, the six most important variables based on PI are retained in training and testing. We anticipate their difference in relative ordering are mostly due to the sample dispersion between the folds. 
Next, we use a more sophisticated method named SHAP \cite{SHAP} for visualizing the feature importance. We also want to compare the feature importance from the trained XGBoost with the ones obtained from SHAP. We use \textbf{Gain} as the default feature importance for the XGBoost model. \textbf{Gain} of a feature is computed by taking its relative contribution in each tree for the whole model. In the comparison between the SHAP feature importance and XGBoost's counterpart, we take the top 20 features in both of them.
In Figure \ref{Figure 1.}, we report the XGBoost feature importance. We compare the same against the mean of the absolute SHAP values presented in Figure \ref{Figure 2.}. Finally, in Figure \ref{Figure 3.} we report the SHAP values of all these features to demonstrate their impact on the model.

It is clear from the comparison that the most important feature (\texttt{married\_1}) is common in both the comparison. Next, \texttt{education.num} is there within the top-3 features in both feature importance ranking. However, the rest of the features including `\texttt{age}' are either not in common or ranked differently based on their importance in the comparisons. 
Next, we move to the summary plots produced by SHAP for a nuanced overview of feature importance.
\subsection{   Summary Plot SHAP}
\begin{figure}
  \includegraphics[width=\textwidth]{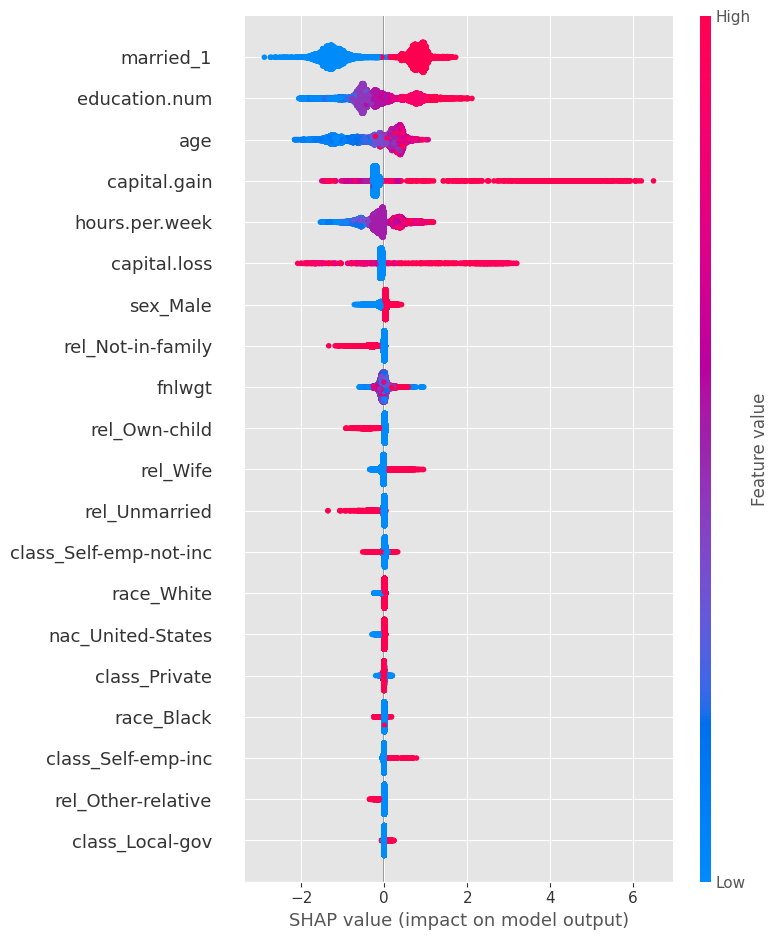}
  \caption{SHAP summary plots}
  \label{Figure 3.}
\end{figure}

The SHAP Summary Plot (Figure \ref{Figure 3.} ) offers more information than the conventional plot. The significance of features is arranged according to decreasing feature relevance.

\textit{Impact on Prediction}: The horizontal axis's position reflects how much or how little the values of the dataset instances for each feature affect the model's output.

\textit{Original Value}: The colour designates a high or low value (within each feature's range) for each characteristic.

\textit{Correlation}: A feature's colour (or range of values) and the effect on the horizontal axis are usually used to assess how well it correlates with the model output. 

Notably, SHAP identifies the variable \texttt{married\_1} as having significant interaction, suggesting that individuals who are married and possess higher educational qualifications are more inclined to earn above \$50,000.


Next, we move to Dependency Plots made by SHAP.
\subsection{Dependency plots}
\begin{figure}
  \includegraphics[width=0.7\textwidth]{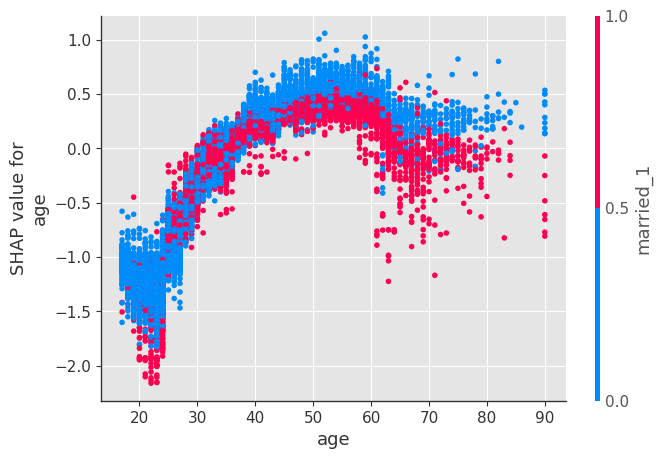}
  \caption{dependency plot: <\textit{age, married\textunderscore1}>}
  \label{Figure 5.}
  \end{figure}

\begin{figure}
  \includegraphics[width=0.7\textwidth]{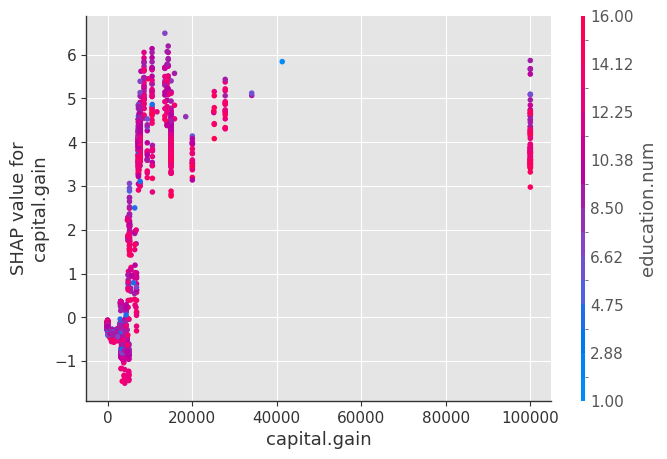}
  \caption{dependency plot: <\textit{capital.gain, education.num}>}
  \label{Figure 6.}
  \end{figure}

\begin{figure}
  \includegraphics[width=0.7\textwidth]{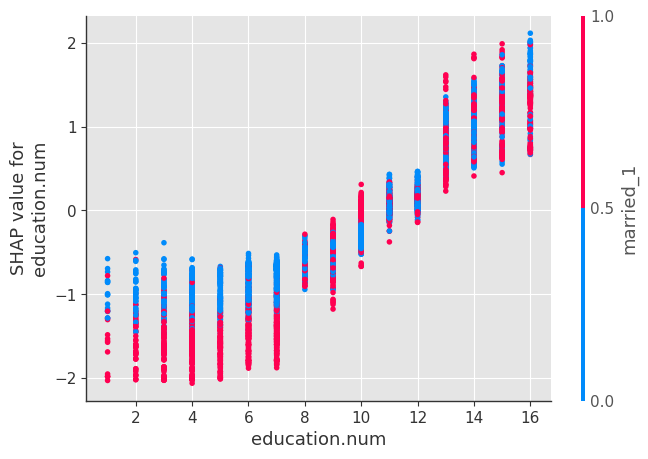}
  \caption{dependency plot: <\textit{education.num, married\textunderscore1}>}
  \label{Figure 7.}
\end{figure}

\begin{figure}
  \includegraphics[width=0.7\textwidth]{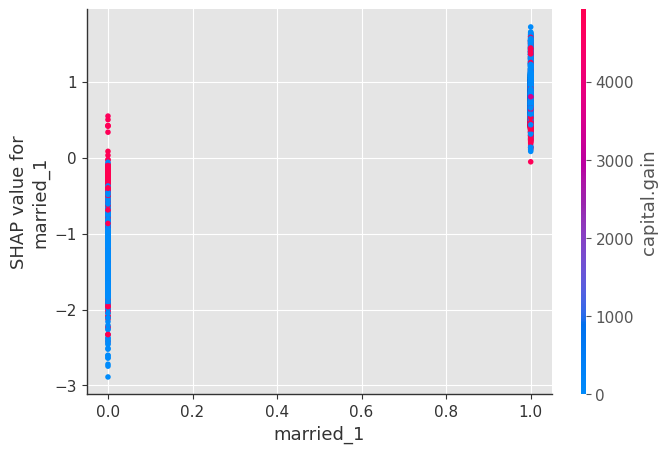}
   \caption{dependency plot: <\textit{married\textunderscore1, capital.gain}>}
  \label{Figure 8.}
 \end{figure}

 \textit{Dependency plot} not only provides the marginal influence the feature has on the model's output, but it also illustrates the relationship with the feature with which it most interacts by colour. This illustrates the model's reliance on the provided feature.  The vertical spread of data points indicates \textit{interaction effects}. We presented some dependency plots (Figure \ref{Figure 5.,Figure 6.,Figure 7.,Figure 8.} with different pairs of features to present how the interaction is between characteristics and how one influences another. 

As depicted in Figure \ref{Figure 5.}, there's a discernible upward trend in marriages between the ages of 20 to nearly 50 years. Meanwhile, Figure \ref{Figure 6.} reveals a pronounced rise in the \textit{education.num} concerning capital gains ranging from 0 to 20k USD. Figure \ref{Figure 7.} further elucidates a consistent progression between educational attainment and marital status. Lastly, Figure \ref{Figure 8.} underscores that married individuals tend to exhibit higher capital gains compared to their unmarried counterparts.

Now, in the next section, we'll discuss post-hoc local explanations using LIME and SHAP.
\section{Discovering Local Scales}
Local surrogate models \cite{surrogate} are interpretable models that are used to explain individual predictions of black-box machine learning models.
\subsection{LIME}
LIME \cite{LIME} stands for Local Interpretable Model-independent Explanations. LIME investigates what occurs in model predictions when the input data is changed. It creates a new dataset containing permuted samples and the old model's related predictions. LIME trains interpretable models (\textit{Logistic Regression, Decision Tree, LASSO Regression, etc}.) on this synthetic set, which are then weighted by the closeness of the sampled examples to the instance of interest.

The explanation, for example, X will be that of the surrogate model that minimises the loss function (performance measure for example, \textit{MSE}- between the surrogate model's forecast and the prediction of the original model), while keeping the model's complexity low.

However, LIME has a distinct problem; it is \textit{inherently} instable. As illustrated in Figure \ref{Figure 10.}, LIME often produces varied explanations for identical instances when executed multiple times. This inconsistency, while not always prevalent, arises from the inherent randomness in generating the surrogate models employed by its linear models. This is not the case with SHAP as it enjoys the uniqueness of SHAP values for a given instance \cite{SHAP}.  There are various ways to visualise SHAP explanations. This chapter will go over two of them: the \textit{Decision \& Force plot} \cite{SHAP}.
\begin{itemize}
\item \textit{Force Plot}:
It shows the effect that each attribute had on the forecast. The output value (model prediction for the instance) and the base value (average prediction for the whole dataset) are the two important numbers to pay attention to. Greater influence is shown by a larger bar, and the colour shows whether the feature value shifted the forecast from the base value to 1 (red) or 0 (blue). 

We run the same experiment $10$ times to show that the explanation is unique and does not exhibit randomness like LIME (Figure \ref{Figure 10.}).
\item \textit{Decision Plots}:
The \textit{Decision Plot: }(Figure \ref{Figure 13.}) displays the same information as the \textit{Force Plot}. The gray vertical line represents the baseline, and the red line indicates whether each characteristic increased the output value higher or lower than the average forecast. Sometimes the information representation depends upon the end user's intention due to which various representations have been engineered under SHAP library. 

\begin{figure}[h!]
  \includegraphics[width=0.450\textwidth]{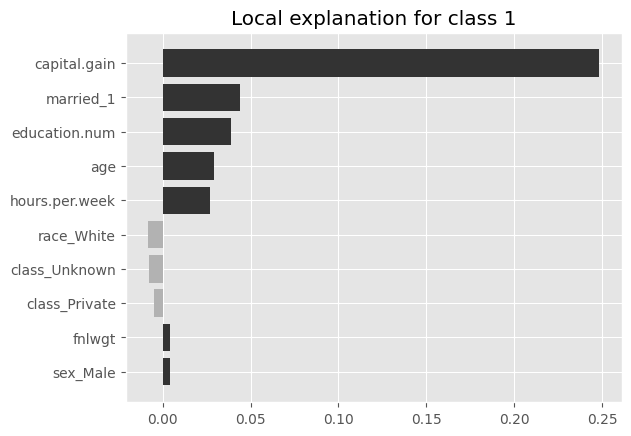}
  \includegraphics[width=0.450\textwidth]{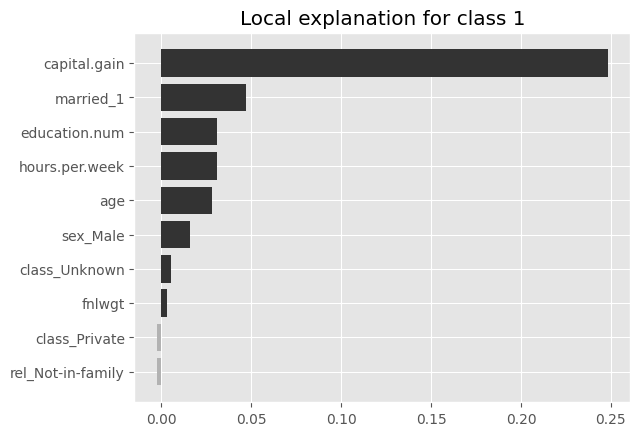}
  \includegraphics[width=0.450\textwidth]{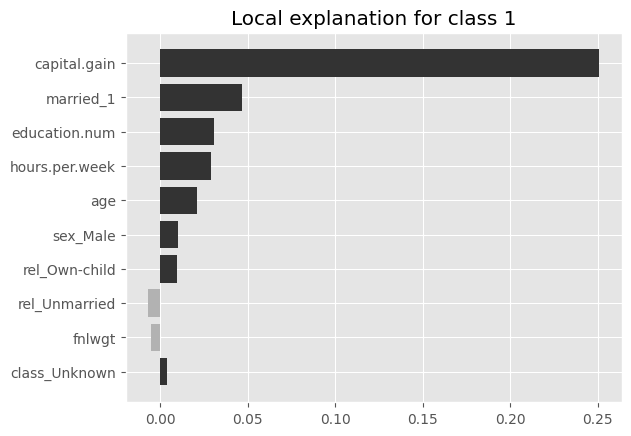}
  \includegraphics[width=0.450\textwidth]{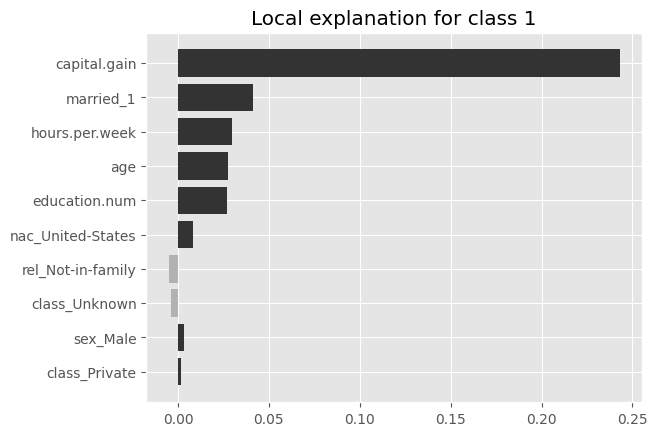}
  \includegraphics[width=0.450\textwidth]{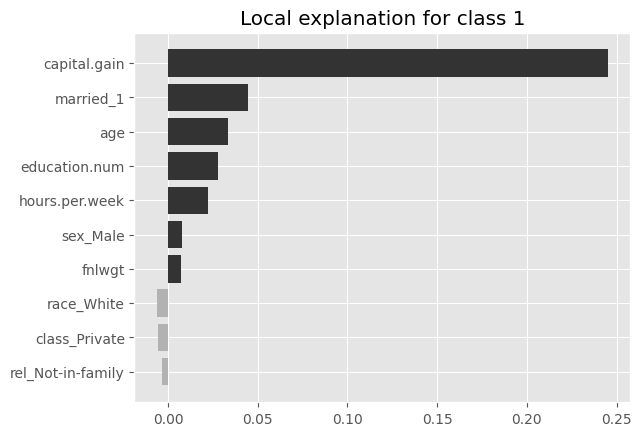}
  \includegraphics[width=0.450\textwidth]{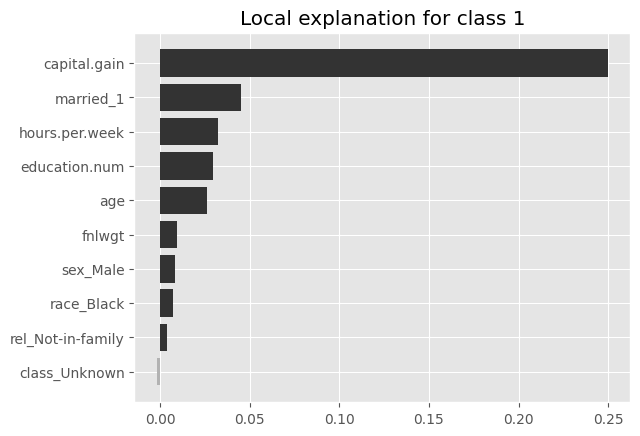}
  \includegraphics[width=0.450\textwidth]{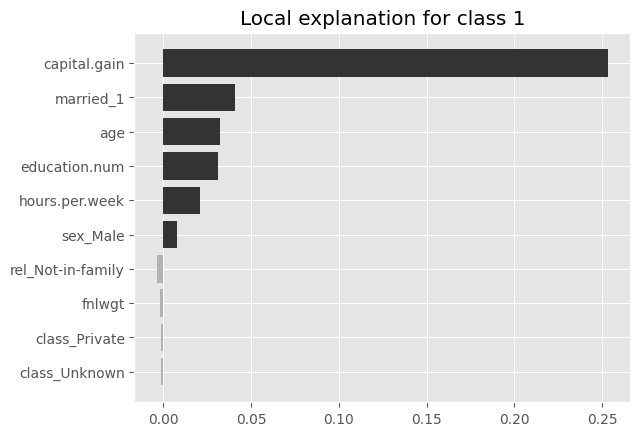}
  \includegraphics[width=0.450\textwidth]{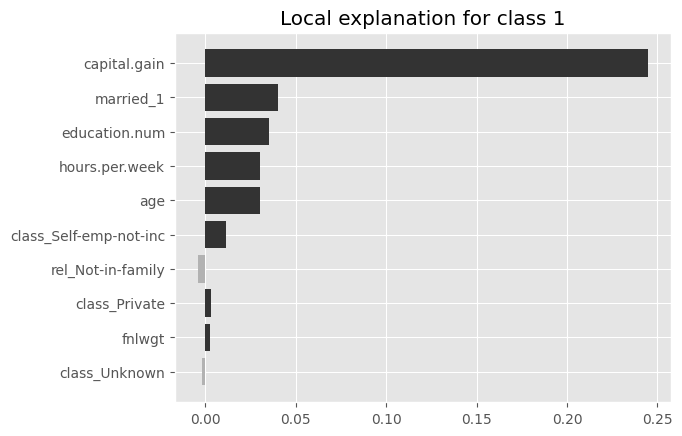}
  \includegraphics[width=0.450\textwidth]{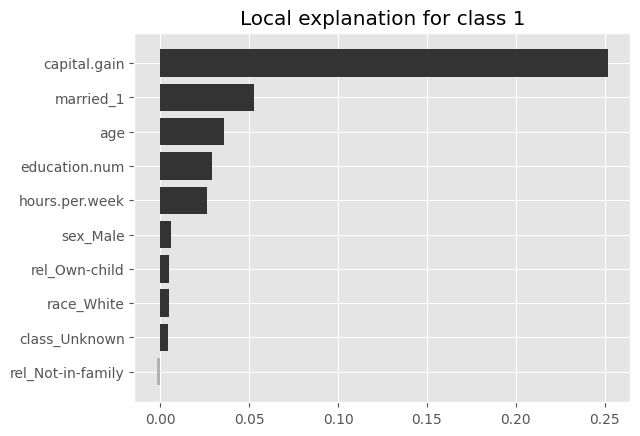}
  \includegraphics[width=0.450\textwidth]{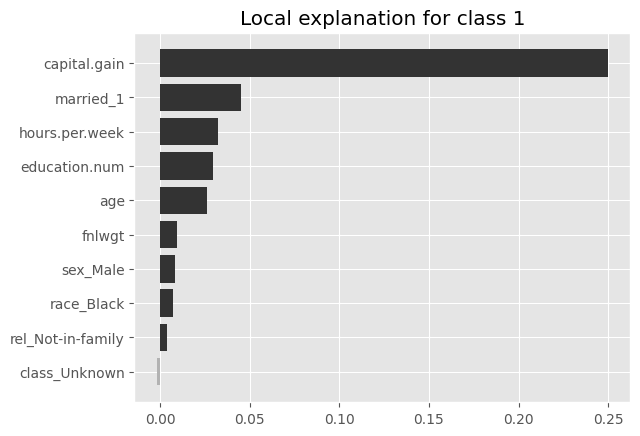}
  \caption{Inherent Instability of LIME}
  \label{Figure 10.}
\end{figure}

\end{itemize}

So far, we have investigated and extensively visualised the feature importance using several representations. Next, we move to mimicking the model globally rather than mere feature importance estimation. We use global surrogate models for globally explaining the model.
\subsection{ Global Explanations}
An interpretable model that has been trained to match the predictions of the black-box model to a very large extent is a global surrogate model. By analysing the surrogate model, we may make inferences about the black box model. Though it, by any means, is not a way to predict the internal architecture of the black box, rather it is meant to deduce some insights regarding the prediction.
Using a \textit{Decision Tree} and a \textit{Logistic Regression} as global surrogate models, we will attempt to approximate the \textit{XGBoost}.\newline
For both the Train and Test sets, R-squared is negative in the case of Logistic Regression (-1.40 , -1.42 respectively). This occurs when the fit is poorer than using the mean alone. As a result, it is argued that Logistic Regression is not an adequate surrogate model.

However, we have obtained $0.77 \& 0.79$ for R-Square scores in train and test respectively in the case of decision tree, so we can proceed with the same if it \textit{mimicks} the trained XGBoost well.

\begin{sidewaysfigure}
    \centering
  \includegraphics[width=\paperwidth]{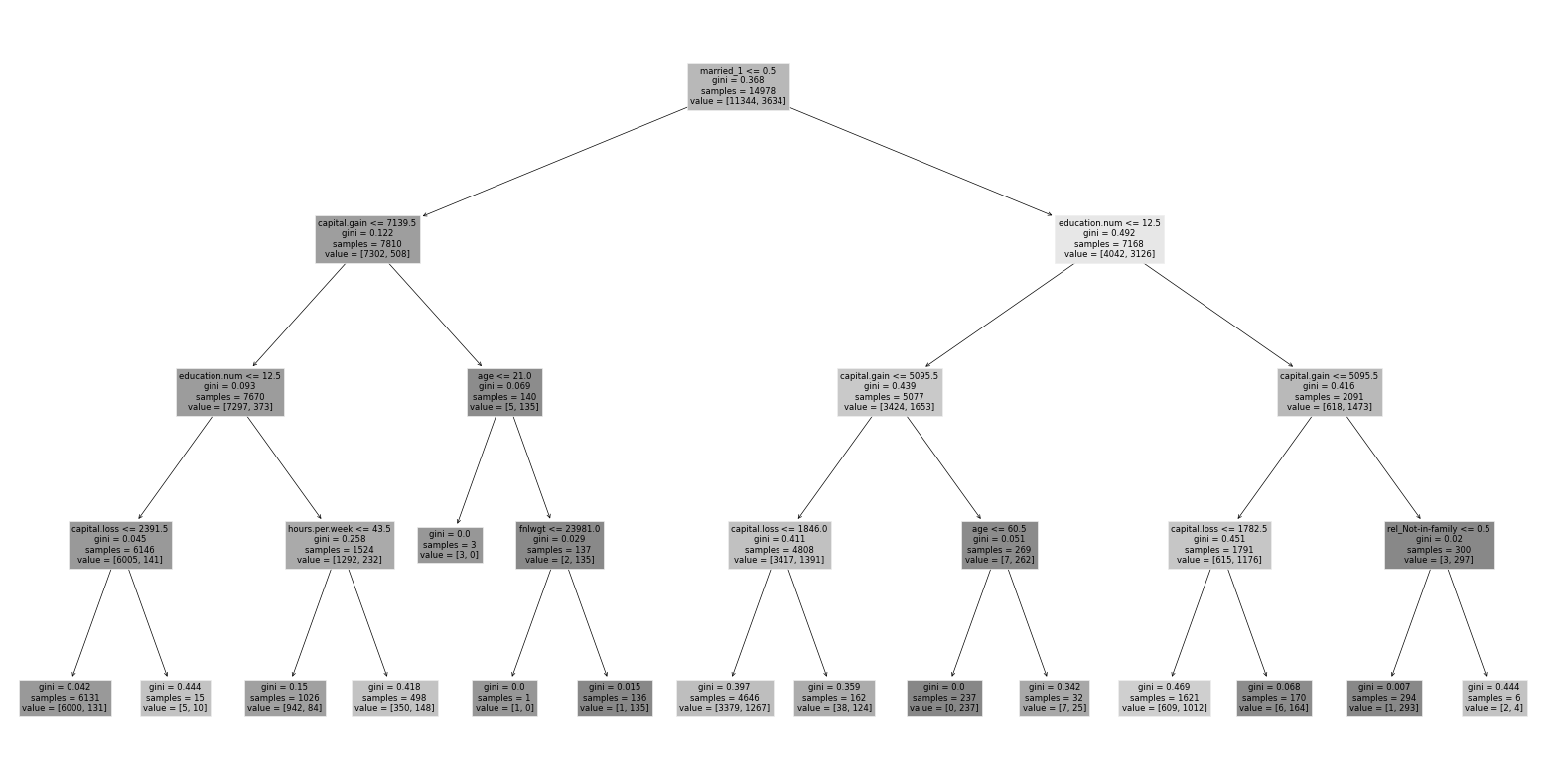}
  \caption{Decision Tree -- mimicking the XGBoost}
  \label{Figure 9.}
\end{sidewaysfigure}

\begin{tabular}{rrrrrr} 
& Accuracy & Precision & Recall & F1 & AUC \\
\hline Train & 0.85 & 0.75 & 0.55 & 0.64 & 0.868197 \\
Test & 0.85 & 0.75 & 0.55 & 0.64 & 0.865062
\end{tabular}\newline
The Decision Tree approximates the \textit{decisions} of the  XGBoost model quite well as per the scores obtained, hence it may be used as a surrogate model for evaluating the main model. However,  it is not guaranteed that the decision tree uses the features in the same way as the XGBoost. It is possible that the tree approximates the XGBoost properly in certain portions of the input space but acts erratically in others as the models are themselves different. We can see in Figure \ref{Figure 9.} that \texttt{married\_1} is the most important feature followed by \texttt{capital.gain} and \texttt{education.num}.  In the importance of the XGBoost feature, these were the top-3 features and in the SHAP dependency plots as well, these were important features to consider. 
Next, we use another method named SP-LIME for generating global explanation.

\subsection{Submodularity:}

\begin{sidewaysfigure}
    \centering
    \includegraphics[width=\paperwidth]{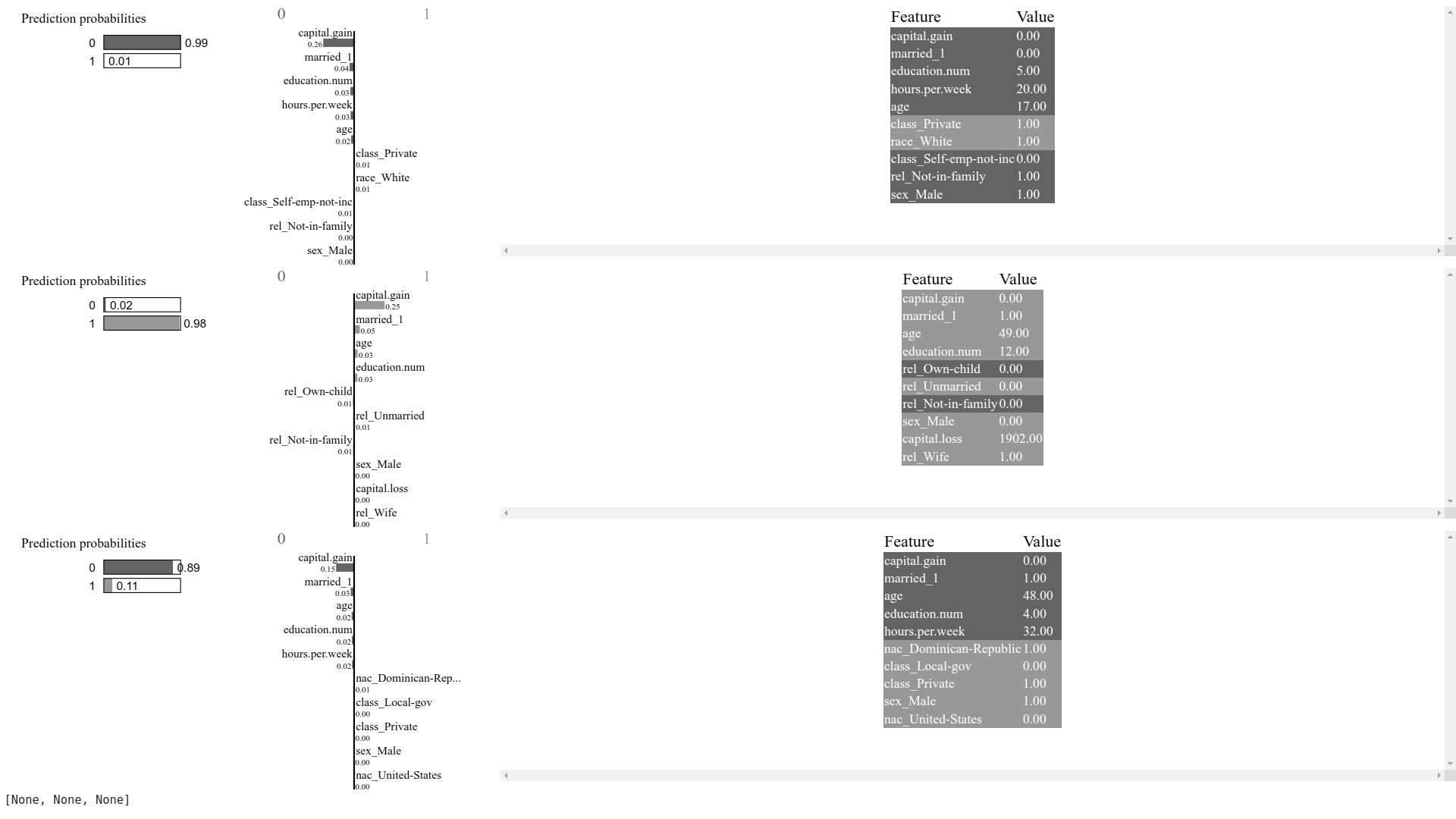}
    \caption{Submodular pic}
    \label{Figure 11}
\end{sidewaysfigure}

 The submodular picking \cite{Submodularity} is important as end users may not have time to examine a large number of explanations. It is critical to choose which instances to explain with care. It seeks to provide explanations for machine learning models by attributing their predictions to human-understandable features. To achieve this objective effectively, it's essential to execute the explanation model across a varied yet representative collection of instances. By doing so, LIME aims to produce a non-redundant set of explanations that collectively offer a comprehensive understanding of the model's behaviour on a global scale. SP LIME algorithm selects a collection of explanations for the user that are diverse and representative, meaning they do not repeat themselves, while also demonstrating how the model works in general as shown in \cite{LIME}. Technically, each instance being a collection of features is a subset of the whole feature set. Collecting a minimal or almost minimal number of subsets to cover the whole or almost the whole set is the main agenda of employing the SP LIME algorithm.
This problem is a derivation of the set cover problem \cite{set cover} which is NP-complete. LIME has its own \textit{greedy picking} \cite{LIME} algorithm which we have demonstrated in Figure \ref{Figure 11}. More technical details can be found at  \cite{LIME}.

In Figure \ref{Figure 11}, $capital.gain$ is shown to be the most important attribute in all individual interpretations that distinguish the classes. Following that, depending on the situation, the most important determinants are $married\textunderscore1$, $education.num$, $age$, and $sex$. These features align with those pinpointed by the algorithms as globally significant.
\begin{figure}[h!]
    \centering
    \includegraphics[width=0.9\textwidth]{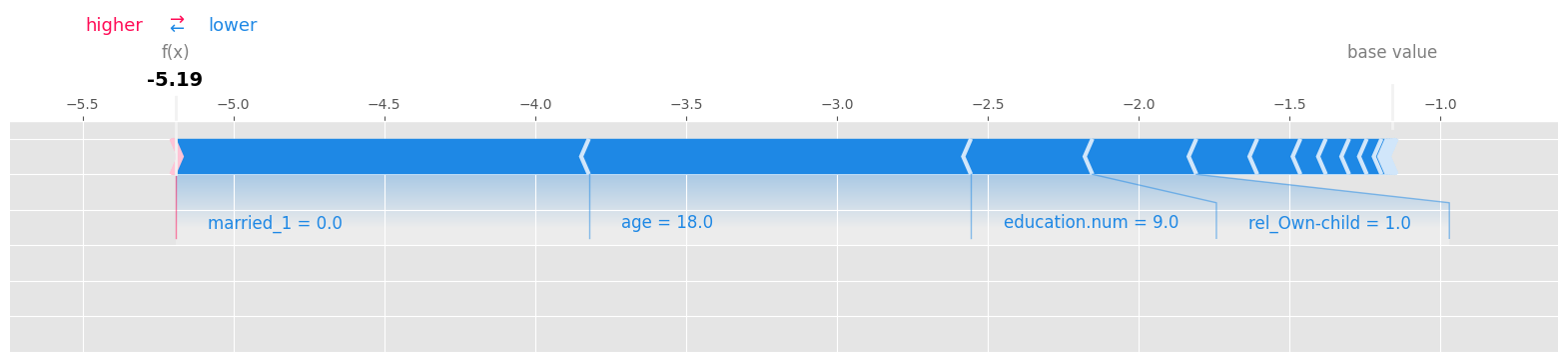}
    \caption{Force Plot}
    \label{Figure 12.}
\end{figure}


\begin{figure}[h!]
    \centering
    \includegraphics[width=\textwidth]{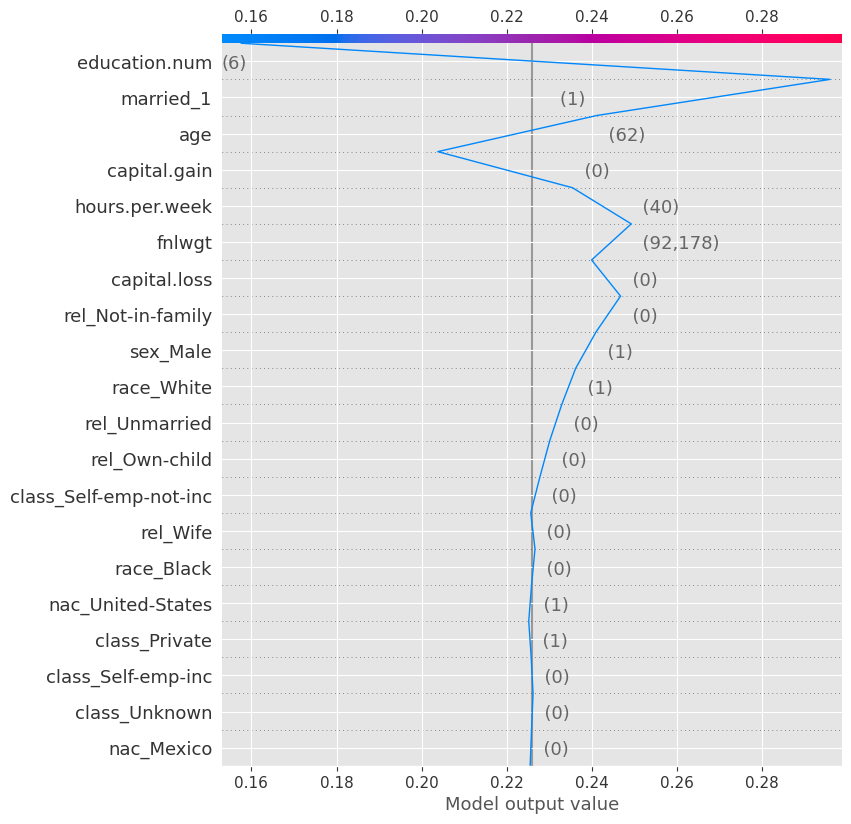}
    \caption{Decision Plot}
    \label{Figure 13.}
\end{figure}

So far, we have investigated how the model's features \textit{interplay} and exhibit patterns. However, whether this \textit{interplay} is \textit{fair \& legitimate} is what we investigate in the next section.
\section{\textit{All} Is Well, but Is \textit{All} \textbf{Fair?}}
In machine learning, "fairness" refers to the several ways that algorithmic bias in machine learning-based automated decision processes is attempted to be corrected. If computer decisions after a machine-learning process were predicated on factors deemed sensitive, they may be viewed as unjust. These variables can be related to gender, ethnicity, sexual orientation, disability, and other factors. Definitions of prejudice and justice are inherently contentious, as is the case with many ethical notions. Fairness and prejudice are typically taken into consideration when making decisions that have an influence on people's lives. The issue of algorithmic bias in machine learning is generally recognised and researched. Many factors have the potential to distort the results, making them unfair to particular persons or groups.

'FairML' \cite{fairML}, the library for fairness in Python helps to investigate these intricacies.

The underlying idea behind FairML (and many other attempts to audit or understand model behaviour) is to change the inputs of a model to quantify its dependency on them. The model is sensitive to a feature if a modest change to an input feature drastically impacts the output.
 In order to quantify how much each characteristic influences the prediction model, FairML projects the input orthogonally.

Let \( F(x, y) \) be a model trained on two characteristics, \( x \) and \( y \). The change in output resulting from a modified input, where \( x' \) perturbs \( x \), can be quantified to measure the model's reliance on \( x \). We express this as:
\begin{equation}
 \Delta F(x', y) = F(x', y) - F(x, y) 
 \end{equation}\myequations{caputing the change in output}
Here, \( \Delta F(x', y) \) represents the change in the model output, and the perturbation in \( x' \) renders the other feature \( y \) orthogonal to \( x' \).
However, linear transformation is an orthogonal projection. FairML uses a basis expansion along with greedy search in the case of non-linear relationships. But, in auditing a model, the determination of fairness is often nuanced and context-dependent. The sensitivity of features can vary across different contexts; what may be considered sensitive in one scenario may not hold the same significance in another. The definition of fairness is multifaceted, influenced by user experience, and cultural, social, historical, political, legal, and ethical considerations. Identifying appropriate fairness criteria involves navigating trade-offs among these factors.
\begin{figure}
    \centering
    \includegraphics[width=\textwidth]{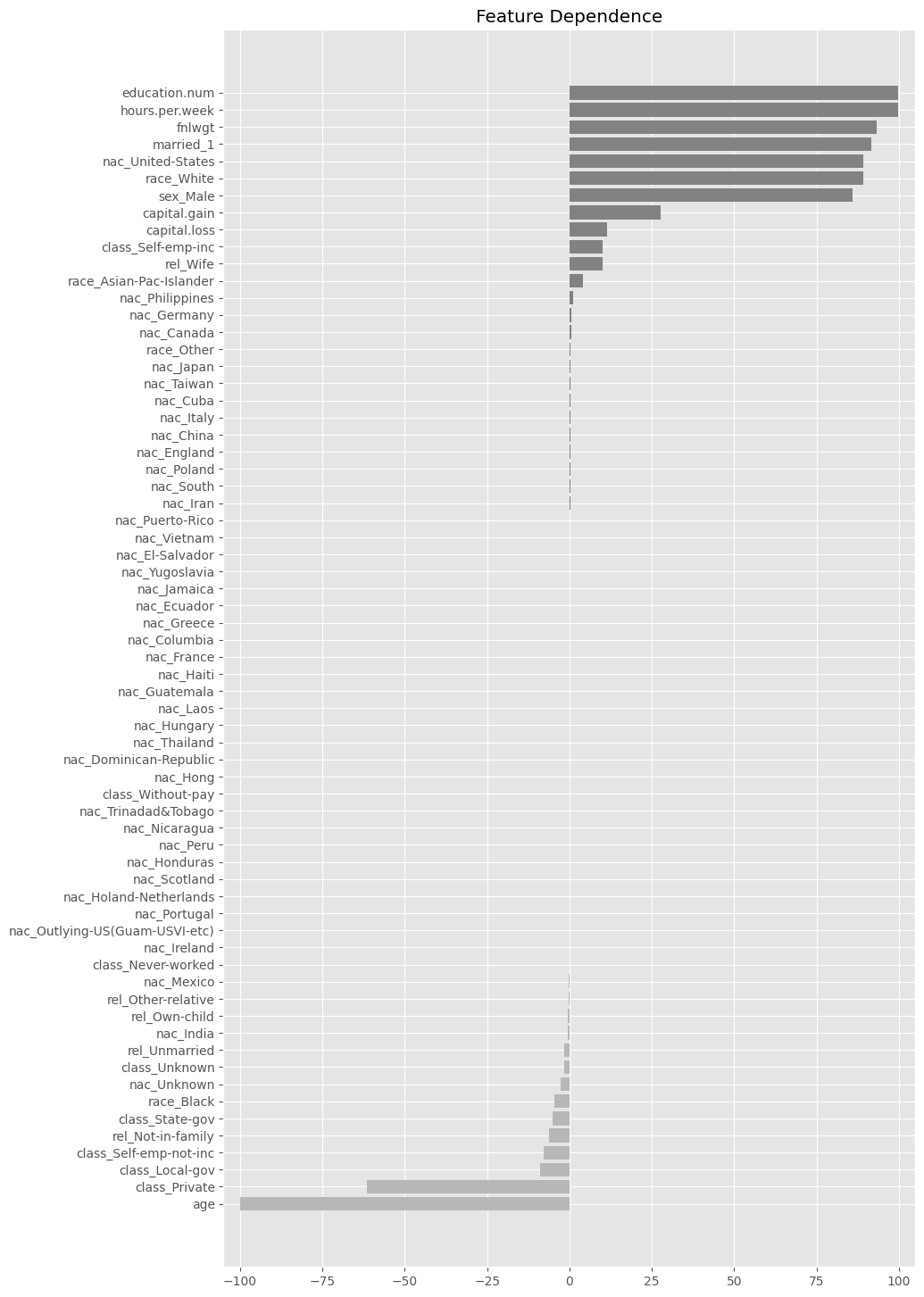}
    \caption{Visualizing Fairness using FairML}
    \label{Figure 14.}
\end{figure}

The visualization: Figure \ref{Figure 14.} uses darker shades to signify a contribution to an output of  Income > USD 50K; lighter shades stand for the opposite. FairML reveals a significant dependence on sensitive features like \texttt{race\textunderscore White}, \texttt{nac\textunderscore United$-$States}, and \texttt{sex\textunderscore Male}. Specifically, the model indicates a strong bias: according to its predictions, an individual is more likely to make more than USD 50k if the individual's nationality is of the United States,  race is white, and sex is male.
Notably, the algorithm's orthogonal projection brings out the relevance of features like \texttt{race\textunderscore White nac\textunderscore United$-$States, \& sex\_Male}, which did not appear in other interpretation methods like SHAP’s Feature Importance. This underscores the importance of investigating for bias along with feature importance. As we observe the popular methods for feature importance may not always be capable of uncovering the intrinsic bias and unfairness in trained models.
Now, in the case of availing \textit{better} education, we started by looking at how parental income influences the education of children and these types of severe bias are what we've found out. This type of societal bias directly impacts the household and the education of the children.
In our assessment, the principles of \textit{fairness and transparency} appear conspicuously absent. \textit{All} is simply, \textit{not} \textbf{Fair} \cite{f1}.
\section{Concluding remark:}


Fairness in machine learning models is a nuanced and evolving concept. Fairness considerations need to account for diverse perspectives, cultural nuances, and the dynamic societal landscape. Moreover, the interpretability of fairness metrics remains an open question. Defining fairness is not a one-size-fits-all endeavour, and achieving fairness often involves navigating complex trade-offs between different dimensions of fairness \cite{fool}.

Our comprehensive analysis delved deeply into the critical interplay between parental income and its consequential impact on educational opportunities, harnessing advanced xAI tools such as LIME, SHAP, and the FairML library. By scrutinizing predictive models derived from the Adult Census data, our deliberate preference for sophisticated algorithms like XG Boost and Decision Trees over simpler models aimed to mirror the inherent complexities typical of real-world black-box systems. However, our findings unveil a disconcerting pattern: pervasive biases persistently permeate these models. While FairML identified these biases as direct orthogonal projections, SHAP analyses sometimes remained oblivious to them. Intriguingly, LIME's volatility introduced yet another layer of complexity, further complicating our interpretative landscape

In conclusion, we want to revisit the objectives: AI came to education for speed and transparency, a strong pillar for availing education is parental income and when that is being scrutinised, it is showing heavy bias under several tests mimicking the real world. The effort we put here to investigate the situation is solely to question the fundamentals: In society, modern education is the pillar to move the nation ahead whereas the \textbf{Scope} of availing education is deliberately getting \textbf{shrunk} due to several biases and education, as the fundamental right to every citizen is being heavily exclusive to a particular direction where the \textbf{Bias} lends itself. Biases present in the training data may be inadvertently perpetuated or even amplified by xAI algorithms, reinforcing existing societal biases \cite{r1,r2,r3}. This is why refining fairness definitions, enhancing the transparency of complex models, and mitigating biases in training data are key challenges that warrant continued attention and collaborative efforts from the research and practitioner communities. Through constructive criticism and iterative refinement, xAI can evolve into a more robust and ethically grounded field \cite{r3,r4}, fostering trust and accountability in the deployment of artificial intelligence systems.
\section{Further Work}
We are consistently striving for improved policies for \textit{Modern} education, emphasizing values of \textbf{Transparency, Equal Opportunity, \& Accessibility}. Our primary focus lies in harnessing the potential of xAI to enhance policymaking. Operating within the realm of public policy, our endeavour centres on exploring innovative avenues to render education policies more accessible, starting from the grassroots level. Leveraging advanced algorithms, we aim to bring about positive transformations that align with our commitment to creating a more transparent, equitable, and accessible educational landscape.\newline


\end{document}